%% file: root.tex
\documentclass[letterpaper, 10 pt, conference]{ieeeconf}  

\IEEEoverridecommandlockouts                              

\usepackage{amsmath,amssymb,amsfonts}
\usepackage{algorithmic}
\usepackage{graphicx}
\usepackage{textcomp}
\usepackage{xcolor}
\usepackage{booktabs}
\usepackage{multirow}
\usepackage{comment}
\usepackage[para]{footmisc}

\usepackage[justification=centering]{caption}
\usepackage{tabularx}

\usepackage[breaklinks,colorlinks]{hyperref}
\usepackage[capitalize]{cleveref}

\title{\LARGE \bf
SPADES: A Realistic Spacecraft Pose Estimation Dataset using Event Sensing
}

\author{Arunkumar Rathinam$^{1}$, Haytam Qadadri$^{2}$ and Djamila Aouada$^{1}$
\thanks{* This work was funded by the Luxembourg National Research Fund (FNR) under the project reference C21/IS/15965298/ELITE. \hspace{2cm} 
}
\thanks{$^{1}$ authors are associated with SnT, University of Luxembourg. {\tt\footnotesize firstname.lastname@uni.lu} }%
\thanks{$^{2}$ author is associated with Télécom Physique, Université de Strasbourg.}%
}

\newcommand{\bm}{\mathbf}
\newcommand{\innerproduct}[2]{\langle #1, #2 \rangle}
\begin{document}

\maketitle
\thispagestyle{empty}
\pagestyle{empty}

\begin{abstract}
In recent years, there has been a growing demand for improved autonomy for in-orbit operations such as rendezvous, docking, and proximity maneuvers, leading to increased interest in employing Deep Learning-based Spacecraft Pose Estimation techniques. However, due to limited access to real target datasets, algorithms are often trained using synthetic data and applied in the real domain, resulting in a performance drop due to the domain gap. State-of-the-art approaches employ Domain Adaptation techniques to mitigate this issue. In the search for viable solutions, event sensing has been explored in the past and shown to reduce the domain gap between simulations and real-world scenarios. Event sensors have made significant advancements in hardware and software in recent years. Moreover, the characteristics of the event sensor offer several advantages in space applications compared to RGB sensors. To facilitate further training and evaluation of DL-based models, we introduce a novel dataset, SPADES, comprising real event data acquired in a controlled laboratory environment and simulated event data using the same camera intrinsics. Furthermore, we propose an effective data filtering method to improve the quality of training data, thus enhancing model performance. Additionally, we introduce an image-based event representation that outperforms existing representations. A multifaceted baseline evaluation was conducted using different event representations, event filtering strategies, and algorithmic frameworks, and the results are summarized. The dataset will be made available at \url{http://cvi2.uni.lu/spades.}
\end{abstract}

\input{sections/00_introduction}
\input{sections/01_related_works}
\input{sections/03_datasets}
\input{sections/04_methods}

\input{sections/05_results}
\input{sections/06_conclusion}





\clearpage

\bibliographystyle{IEEEtran}
\bibliography{references}

\end{document}

%% file: sections/00_introduction.tex
\section{Introduction}
\label{sec:intro}

The rise of Deep Learning (DL) algorithms motivated state-of-the-art spacecraft pose estimation methods to leverage deep neural networks (DNN) to infer the pose of a known non-cooperative spacecraft from a single RGB image \cite{sharma2018pose, proenca2020, chen2019}. However, they require abundant labeled data for training, while acquiring orbital imagery data is expensive and challenging, considering each target is unique. Many satellite pose estimation models are trained using synthetic images to overcome this limitation. However, this leads to the \textit{domain gap} or \textit{Sim2Real} problem \cite{park2022speed+}, that is, models trained in one domain (synthetic) face a drop in performance when tested in another (real data) due to overfitting of the features specific to the training domain. To alleviate this problem, \textit{Domain Adaptation} (DA) methods \cite{park2021} are adopted to increase the performance of the model in the target domain, using techniques such as adversarial learning and reconstruction approaches.
\begin{figure}[ht]
    \centering
    \includegraphics[width=\linewidth]{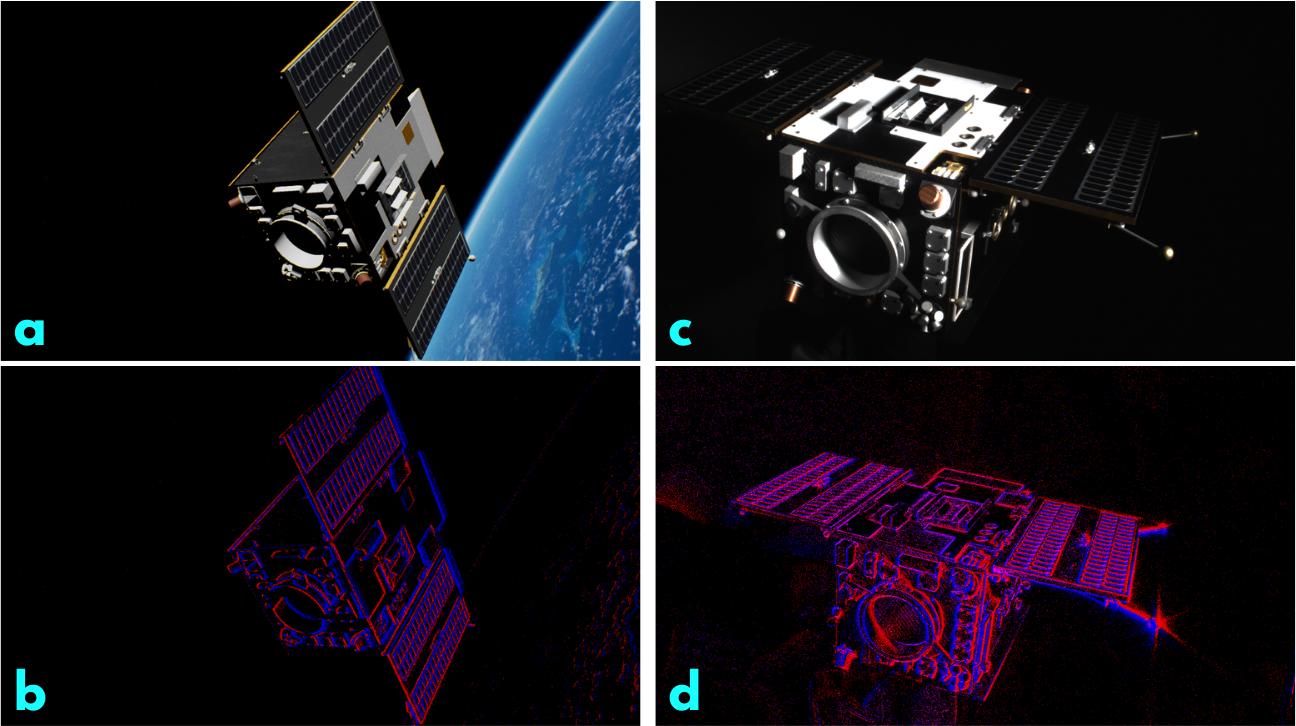}
    \caption{Samples from the SPADES dataset. (a) RGB image generated using Unreal Engine, (b) Event data generated using the ICNS simulator, (c) Real image acquired in the lab, (d) Real event data acquired in the lab.}
    \label{fig:overview-dataset}
    \vspace{-2em}
\end{figure}

Event sensing was proposed in \cite{jawaid2023} as a solution to reduce the domain gap for DL-based spacecraft pose estimation without requiring DA techniques. Indeed, Event cameras have gained attention in space applications \cite{cohen2019event, afshar2020} due to its potential benefits. These sensors capture sparse data, and each pixel is independently activated by changes in light intensity, leading to asynchronous responses. Notable advantages include high temporal resolution (up to 1$\mu$s), a wide High Dynamic Range (HDR) (typically up to 140 dB), low latency, and low power consumption \cite{gallego2022}. Their higher HDR values result in smaller solar exclusion angles, making them well-suited for orbital sensing. Event sensors' HDR and asynchronous response characteristics help perceive the target in a way that reduces sensitivity to drastic illuminations, thus narrowing the domain difference \cite{jawaid2023}.

The SEENIC dataset \cite{elms2022seenic} proposed in \cite{jawaid2023} was the first and only event sensing dataset available for spacecraft pose estimation tasks. The advantages and limitations of this dataset are briefly discussed in \cref{sec:rel-works}. Building on the work in \cite{jawaid2023} and aiming to gain a deeper understanding of the behavior of the event data in more realistic orbital scenarios while facilitating the training and evaluation of the DL models, we introduce a \textbf{novel event dataset}, called \textbf{SPADES} - \textit{\textbf{SPA}cecraft Pose Estimation \textbf{D}ataset using \textbf{E}vent \textbf{S}ensing}, as our \textit{first} contribution. The proposed SPADES dataset employs the Proba-2 satellite of the PROBA-2 mission \cite{dlrproba2} as a target. This dataset comprises simulated event data and real event data collected using a realistic satellite mockup and the SnT Zero-G testbed facility \cite{pauly2022lessons}. Our \textit{second} contribution involves a \textbf{novel event frame filtering} approach, a data pre-processing technique that selects event frames with sufficient shape information for training, thereby assisting the model in better learning and improved performance. As our \textit{third} contribution, we introduce a \textbf{image-based event representation} with three channels designed to take advantage of existing 2D convolutional neural networks (CNN) while providing superior performance compared to existing representations. Finally, to assess baseline performance, we implement existing DL algorithms on two prominent spacecraft pose estimation approaches and present the results.

The paper is arranged as follows: \cref{sec:rel-works} presents related datasets, algorithms, and event representations. \cref{sec:syn-dataset} presents the proposed SPADES dataset, including synthetic data generation pipeline and real data acquisition. \cref{sec:methods} introduces the new event representation and filtering approach. \cref{sec:results} presents the evaluation results and finally \cref{sec:conclusion} presents the conclusion.

%% file: sections/01_related_works.tex
\section{Related Works}
\label{sec:rel-works}

\subsection{Datasets for Spacecraft Pose Estimation}
\label{sec:dataset_spe}

The first generation image datasets, SPEED \cite{speed2019} and URSO \cite{proenca2020}, were oriented toward synthetic data and used simulators to generate realistic renderings of targets in orbit. Recent image datasets, such as SPEED+ \cite{park2022speed+}, SPARK 2022 \cite{rathinam2022dataset}, and SHIRT \cite{park2022shirt}, have included real data from laboratory setups in addition to simulated data. This inclusion serves to validate the performance of the DL algorithms in realistic scenarios that simulate space environments.

An event-sensing dataset, SEENIC \cite{elms2022seenic}, was introduced in \cite{jawaid2023} to assess the domain gap in spacecraft pose estimation using event data. Despite being the first and only event dataset for spacecraft pose estimation, there are several limitations associated with this dataset, which are summarized below. First, the target model is the Hubble Space Telescope (HST), whose actual dimensions are 13 m in length and 4 m in width. A scaled version of the HST (approximately 1:40 to 50) was used for real data collection, and the precise dimensions of the physical and simulation models were not disclosed. Second, the HST mockup was 3D printed for simplicity, lacking precision and surface texture, which affected the overall quality of the dataset. Third, the real event data lack relative pose labels for the target in the camera reference frame, and the authors rely on measurements between successive poses for their metrics. Finally, the synthetic dataset was generated from a single trajectory, resulting in an imbalance in the pose distribution, and also lacks variation in lighting scenarios.

\begin{figure*}[t]
    \vspace{0.5em}
    \centering  
    \includegraphics[width=0.95\linewidth]{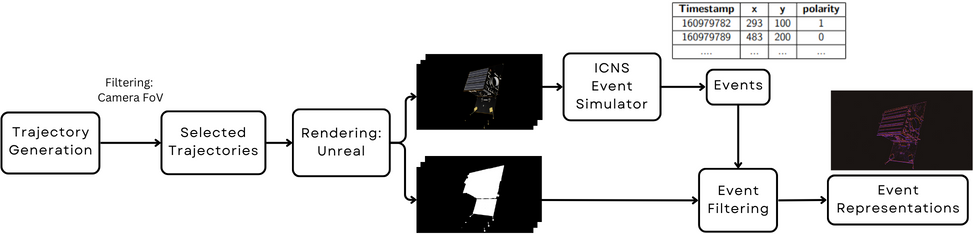}
    \caption{Overview of the synthetic data generation pipeline.}
    \label{fig:synthetic-gen-pipeline}
    \vspace{-1.5em}
\end{figure*}

\subsection{Algorithms}

The two prominent approaches to estimating satellite pose based on DL are the \textit{ Direct or End-to-End} approach and the \textit{Hybrid Modular} approach \cite{rathinam2021, pauly2023survey}. The direct approach \cite{proenca2020, sharma2018pose} is based on direct regression of pose labels from images, while the hybrid pipeline \cite{chen2019} involves a sequence of steps. This includes using an object detection network to detect the target, followed by a keypoint regression network to regress the 2D keypoints location, and finally, utilizing the Perspective-n-Point (PnP) solver to estimate the pose from 2D-3D correspondences. A summary of DL-based satellite pose estimation approaches is provided in \cite{pauly2023survey}.

The baseline evaluation on the SEENIC dataset in \cite{jawaid2023} employed a \textit{Hybrid} pipeline (without DA techniques) trained with synthetic data and tested with real data. During the performance evaluation on real data, the authors resort to measuring errors between successive poses as a performance metric due to hardware constraints that prevented them from directly obtaining the true relative pose of the object within the camera reference frame. However, it should be noted that such metrics are susceptible to errors that accumulate over time due to drift. Even though errors between successive poses may initially appear minor, they can eventually lead to a significant deviation from the actual ground truth.

To mitigate such issues and assess performance using standard pose metrics, the proposed SPADES dataset is supplemented with ground-truth pose labels containing target poses in the camera reference frame for both data modalities.

\subsection{Event Data Processing}

An event stream is the sequence of events triggered due to the change in light intensity as recorded by individual sensor pixels. Each event readout in the form of a tuple $e = (x,y,p,t)$, where $x$ and $y$ denote the pixel coordinates, $p$ indicates an increase or decrease in intensity (polarity), and $t$ represents the global timestamp of the event in microseconds ($\mu$s) as recorded in the camera timeline. A sequence of events over a time window of $\tau$ can thus be represented as $ E_{\tau} = \{ e_i\ | \ t<i<(t + \tau) \} $. These accumulated events can be processed and represented in various formats, including images \cite{jawaid2023, rudnev2020, lagroce2017}, voxels \cite{xie2022voxel}, graphs \cite{bi2020graph}, 3D point sets \cite{sekikawa2019eventnet}, and motion-compensated event images \cite{gallego2017acc}.

\subsubsection*{Image-based Representations} The image-based representations convert sparse events to dense frames to leverage existing CNN architectures. Event-to-Frame (E2F) \cite{jawaid2023} representation works by accumulating events over a given time window or an event batch, followed by normalization and exported as an intensity image. The Locally-Normalised Event Surfaces (LNES) \cite{rudnev2020} representation effectively retains temporal and polarity information during the conversion. Within LNES representation, each event frame consists of two channels, $I \in \Re ^{W\times H\times 2}$, distinguished by the polarity of the event. Using individual channels for positive and negative events preserves the polarities and limits event overriding \cite{rudnev2020}. The Time Surfaces (TS) \cite{lagroce2017} representation aims to preserve temporal information from the event stream while discarding polarity details.  Unlike LNES, TS is generated by applying an exponential decay to the time within the time window using the last set of events recorded in the neighbourhood of the current event $e_i (x,y)$. 

%% file: sections/03_datasets.tex
\section{Dataset}
\label{sec:syn-dataset}

\subsection{Synthetic Data Generation}
\subsubsection*{Trajectory Selection}

The process of generating an event stream involves creating a sequence of images. This can be achieved by moving either the camera or the target while keeping the target within the camera's field of view (FoV). Our data generation pipeline employs a fixed camera and a moving target. The trajectory generation comprises two steps.

The \textit{first step} is to initialize the starting and ending poses of the sequence, denoted as $[\mathbf{q_{start}}|\mathbf{t_{start}}]$ and $[\mathbf{q_{end}}|\mathbf{t_{end}}]$, where $\mathbf{q_{x}}$ represents the orientation as quaternions and $\mathbf{t_{x}}$ represents the positions as a translation vector. Quaternions were sampled from a uniform distribution. In the translation vector, $t_z$ ranges between 3.5 and 12 m, determined based on factors such as focal length, sensor size, resolution, and target size; $t_x$ and $t_y$ are constrained by the camera's FoV. 

The \textit{second step} involves interpolation between the start and end pose over $n$ steps: $S = ([\mathbf{q_{0}}|\mathbf{t_{0}}],[\mathbf{q_{1}}|\mathbf{t_{1}}],...,[\mathbf{q_{n}}|\mathbf{t_{n}]})$.  The interpolation methods employed are either Helix or Spline interpolation. After generating each sequence, each pose within the sequence is verified with 2D keypoint projection results, ensuring that all edge keypoints remain within the image. Table \ref{tab:syn-dataset-char} summarizes the size of the dataset, the number of trajectories, the interpolation methods, and the characteristics of the range.  \cref{fig:synthetic-gen-pipeline} illustrates the complete data generation pipeline.

\vspace{0.5em}
\subsubsection*{RGB data}
After generating the ground truth sequence, we render RGB images using a Unreal Engine\footnote{www.unrealengine.com} (UE) simulator. To render these synthetic images, we use the CAD model of the Proba-2 satellite downloaded from the ESA Science Satellite Fleet\footnote{http://scifleet.esa.int}. Communication with the UE environment is facilitated through the UnrealCV library \cite{qiu2016}. The UE environment incorporates 16k Earth texture maps from the Blue Marble collection\footnote{visibleearth.nasa.gov}, employs physically-based shading, and includes Rayleigh scattering to simulate atmospheres. Prior to rendering, camera poses are randomly sampled and fixed for each sequence, resulting in diverse backgrounds and lighting scenarios. The target is placed relative to the camera pose using the corresponding ground-truth pose, and the images are subsequently rendered.


\vspace{0.5em}
\subsubsection*{Event data}
\begin{figure}[ht]
    \centering
    \includegraphics[width=0.9\linewidth]{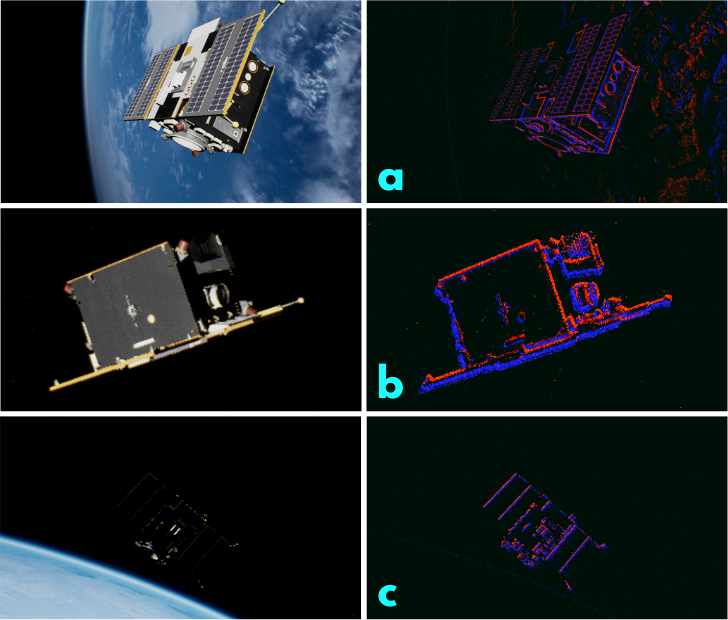}
    \caption{Synthetic data samples from RGB and Event sensor. (a) images with good lighting and background, (b) images with good lighting and no background, and (c) images with harsh lighting.}
    \label{fig:syn-samples}
    \vspace{-1.5em}
\end{figure}

The event data stream is generated using the ICNS event simulator \cite{joubert2021}, which uses Blender\footnote{www.blender.org} to simulate the behavior of neuromorphic sensors. This simulator offers a more realistic simulation of the sensor output by accurately modeling the sensors' pixel-level behavior, taking into account factors such as latency, noise, and other relevant characteristics. Samples from generated synthetic event data are depicted in \cref{fig:syn-samples}.

\subsection{Real Data Collection}
\label{sec:real-dataset}

\subsubsection*{Testbed}
The Zero-G Laboratory facility at the Interdisciplinary Centre for Security, Reliability and Trust (SnT) in the University of Luxembourg \cite{pauly2022lessons} was used for real data acquisition. The laboratory setup covers a space with dimensions of $5\times 3\times 2.3$m (WxLxH) and is equipped with two UR10 robotic arms mounted on two linear rails, one on the side wall and the other on the ceiling. Furthermore, the facility is equipped with an OptiTrack motion capture system (OTS) comprising eight cameras that enable tracking of a predefined rigid body fitted with either active or passive markers.
\vspace{0.5em}
\subsubsection*{Event Camera} 
The camera utilized in the data acquisition is Prophesee Metavision EVK4-HD\cite{prophesee2023} equipped with the SONY IMX636ES(HD) event vision sensor, representing the latest technology available at the time. This camera boasts a resolution of $1280 \times 720$ pixels on a $1/2.5"$ sensor, each pixel measuring $4.86 \mu$m. Additionally, it is equipped with a 6mm fixed focal length lens, providing a horizontal FoV of 54.6\textdegree. The maximum read-out throughput is 3 Gevents/s, and the typical power consumption is 0.5W to 1.5W max. 

\vspace{0.5em}
\subsubsection*{Proba-2 Mockup}
The satellite mockup used in the experiments weighs approximately 7 kg, with a scaling ratio of 1:2.5. The dimensions of the physical model, along the X, Y and Z axes, are $0.64\times 0.24\times 0.416$m, respectively. The mock-up was manufactured through a third-party vendor, and the materials were carefully selected to minimise deviations from the textures found in the CAD data.

\begin{figure}[ht]
    \centering
    \includegraphics[width=0.7\linewidth]{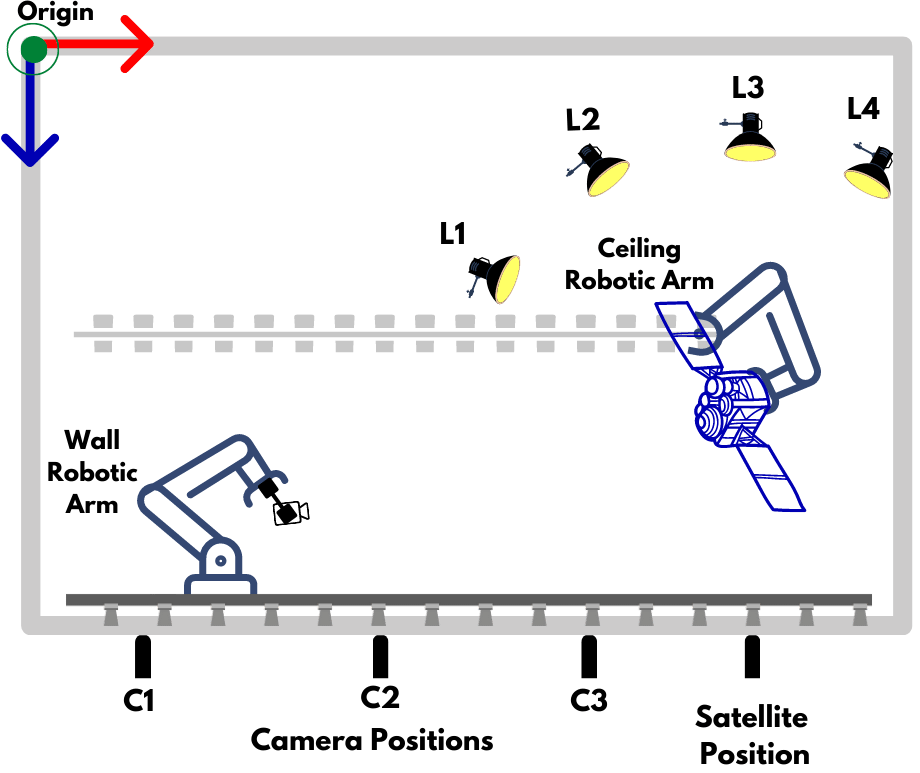}
    \caption{Schematic of the Zero-G lab setup for real data collection.}
    \label{fig:zero-g-layout}
    \vspace{-1em}
\end{figure}

\vspace{0.5em}
\subsubsection*{Light Setup} 
The intensity of sunlight in orbit corresponds to the solar irradiance of 1366 W/m$^2$ or illuminance of $\sim$ 163,000 lux. To emulate orbital lighting, the Aputure LS-600D-PRO LED lamp was used as a light source for data collection. The lamp can produce 224,200 lux at a distance of 1 m when mounted with a Fresnel F10 lens with a spot angle of 15\textdegree. In our setup, the lamp is fixed at a distance of 1.5 m as a trade-off between safety and
accuracy, producing 120,000 lux for the color temperature 5800K.

\begin{figure}[ht]
    \centering
    \includegraphics[width=0.9\linewidth]{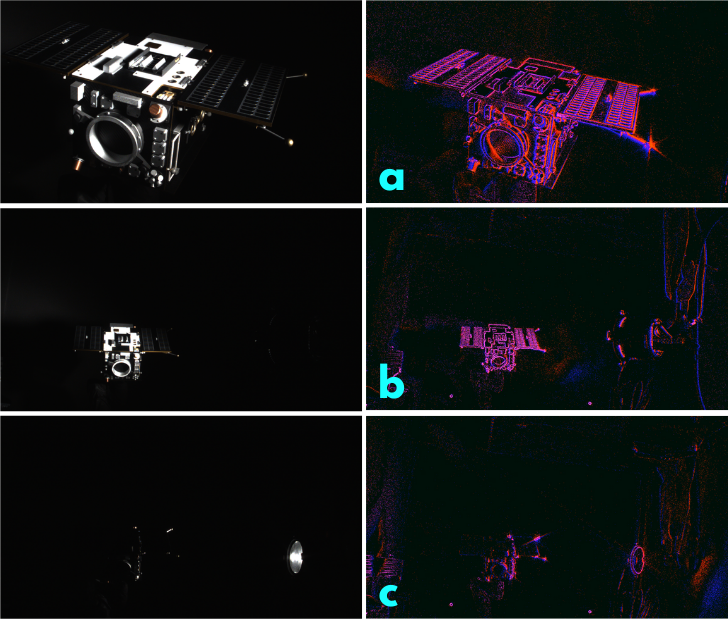}
    \caption{Real data samples from FLIR RGB camera and Prophesee Event Camera. (a) Close-range with L1, L2 lighting, (b) Far-range with L1, L2 lighting, and (c) Far-range with L3, L4 lighting. }
    \label{fig:real-samples}
\end{figure}

\subsubsection*{Event camera calibration}
To calibrate the event camera, we used grayscale image reconstruction \cite{muglikar2021}, which leverages a neural network-based image reconstruction technique to move from events to grayscale image. The camera is moved around the fixed calibration board to collect the calibration sequence. The event stream is extracted into batches with a fixed time window during processing. Grayscale image reconstruction for each batch of events was achieved using the E2VID model \cite{Rebecq19pami}, and the corresponding pose labels were extracted from the OTS. The reconstructed images were subsequently processed to compute the camera intrinsics using the MATLAB Camera Calibration toolbox. For extrinsic calibration, the hand-eye calibration approach \cite{radu_handeye_1995} was used to find the transformation between the actual camera reference frame and the rigid body camera frame defined in OTS. This fixed transformation maps the raw pose label of the rigid body to the actual camera pose in the OTS coordinate system. Similarly, the satellite mockup has a pre-defined marker setup to collect pose labels in the OTS coordinate system. Data between the actual camera reference frame and satellite poses were synchronized on the basis of timestamps, and the transformation was applied to yield ground truth data representing the relative pose information of the object in the camera reference frame. 

\vspace{0.5em}
\subsubsection*{Data Collection}

During real data acquisition, various combinations of lighting conditions (L1, L2, L3, L4) and camera positions (C1, C2, C3) were employed, as illustrated in \cref{fig:zero-g-layout}. Based on the camera's motion, the trajectories are categorized into two groups: \textit{static} and \textit{dynamic}. In \textit{static} trajectories, the camera (representing the chaser satellite) maintains a constant distance from the target, observing the target's motion, thereby emulating the observation phase. In \textit{dynamic} trajectories, the camera approaches the target with linear or spiral motion, while the target exhibits stationary or rotational movement along its Y-axis. Further details on the real data set can be found in Table \ref{tab:syn-dataset-char}.

\vspace{-0.5em}
\begin{table}[ht]
\centering
\resizebox{\linewidth}{!}{%
\begin{tabular}{lll}
\toprule
 & \textbf{Synthetic} & \textbf{Real} \\ 
\midrule
\textbf{Sensor resolution} & 1280x720 & 1280x720 \\ 
\textbf{Dataset size} &  179,400 (no. of poses)  & 15,500\\ 
\textbf{No. Trajectories} & 300 & 31 \\
\textbf{No. poses/traj} & 598 & 500\\ 
\textbf{Interpolation}  & 80\% spline \& 20\% Helix & -\\ 
\textbf{Range}  & 3.5 - 12 m & 3.5 - 9 m\\ 
\textbf{Range dist.}  & Close, Mid, Far, Limit & Close, Mid, Far\\ 
\textbf{Lighting}  & Easy, Hard & L1, L2, L3, L4\\ 
\textbf{Rendering} & Unreal Engine (RGB) & -\\ 
\textbf{Event Camera} & ICNS Emulator & Prop. EVK4HD \\
\textbf{Background} & Earth & -\\ 
\textbf{Filtering} & Bbox/Mask & Min. event count \\ 
\bottomrule
\end{tabular}}
\caption{Summary of proposed SPADES dataset}
\label{tab:syn-dataset-char}
\vspace{-1.5em}
\end{table}

%% file: sections/04_methods.tex
\section{Event Data Processing}
\label{sec:methods}

\subsection{Event Representation: 3-Channel}
\begin{figure}[ht]
    \centering
    \includegraphics[width=\linewidth]{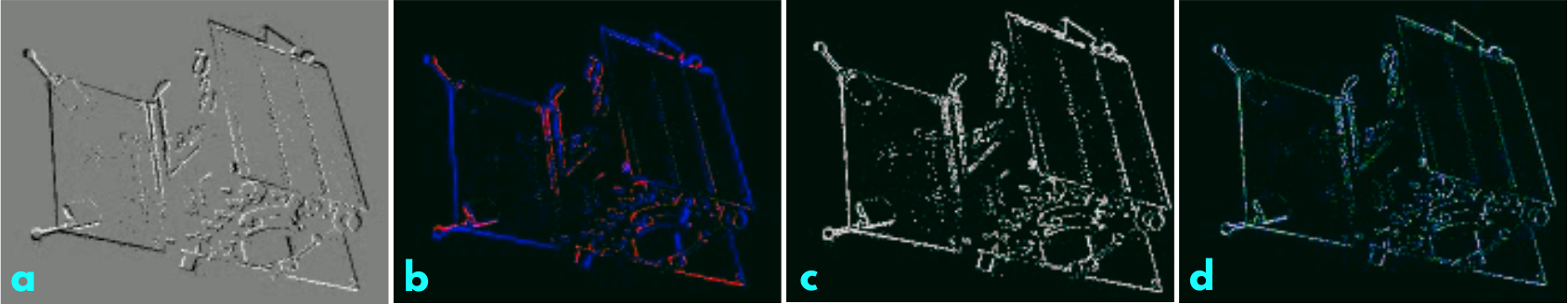}
    \caption{Event representations (a) E2F, (b) LNES, (c) TS, (d) 3C}
    \label{fig:diff-event-repr}
    \vspace{-1.5em}
\end{figure}

The proposed event representation, namely 3-Channel (3C), is a pseudo frame with three channels to leverage the algorithms designed for 3-channel RGB images. Improving upon the TS representation \cite{lagroce2017}, the 3C Representation uses exponential decay to track temporal information while it splits the actual time window \textit{W} into three sub-windows of size \textit{W/3}. Events collected within each sub-window are processed independently and organized into channels in chronological order. This approach ensures the segregation of maximal temporal information into separate channels, preserving the inherent asynchronous nature of events by dividing the time window into sub-windows. This representation differs from a three-channel presented in \cite{bai_accurate_2022}, which uses the polarity information and event count. Different event representations are presented in \cref{fig:diff-event-repr} for comparison.

\subsection{Event Frame Filtering}
\label{sub-sec:data-filt}
The preliminary analysis of the synthetic data suggested that not all generated event frames are suitable for training. As the satellite rotates, the change in the incidence angle of light may result in fewer events generated. This renders some of the frames useless to recover any information. Therefore, there is a need to filter good quality event data to improve learning and eventually model performance. To achieve this, we propose a novel \textit{mask-based filtering} approach to filter good event frames using the segmented mask of the target within the image. First, we define a discrete uniform distribution $\mathbf{p_{\text{uniform}}}$, as follows,
\begin{equation}
    \begin{array}{ccccc}
    \mathbf{p_{\text{uniform}}} & : & \mathbf{M_{\text{pixel}}} & \to & \mathbb{R} \\
     & & (x, y) & \mapsto & \frac{1}{N} \\
    \end{array}
\end{equation}
where $N$ is the number of pixels of the mask and $\mathbf{M_{\text{pixel}}}$ is the set of pixels (x, y) within the mask. Next, a discrete distribution for event data $\mathbf{p_{\text{event}}}$, as follows,
\begin{equation}
    \begin{array}{ccccc}
    \mathbf{p_{\text{event}}} & : & \mathbf{M_{\text{pixel}}} & \to & \mathbb{R} \\
     & & (x, y) & \mapsto & \begin{cases} 
                              \frac{0.99}{N} & \text{if } (x, y) \in \mathbf{E} \\
                              \frac{0.01}{N} & \text{if } (x, y) \notin \mathbf{E} \\
                           \end{cases}
    \end{array}
\end{equation}
where $\mathbf{E}$ is the event stream. Subsequently, the KL-divergence is calculated between the two $\mathbf{p_{\text{event}}}$ and $\mathbf{p_{\text{uniform}}}$ to filter out inadequate pose labels using a threshold set across the dataset. For comparison, a direct filtering method is presented, \textit{ bbox-based filtering}, using the bounding boxes of the objects. In this method, the ratio of events count within a given bounding box to the area of the bounding box is computed, and a pre-defined value is set as a threshold to exclude the pose labels inadequate for training.

\begin{figure}[ht]
    \centering
    \includegraphics[width=\linewidth]{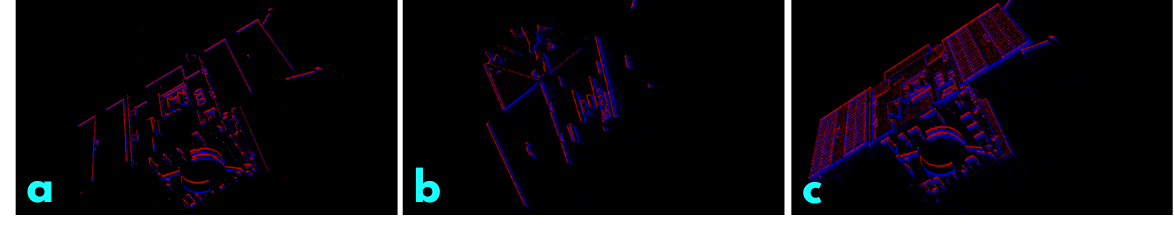}
    \caption{Mask-filtering samples. (a) and (b) samples removed from training data, (c) samples included in training data.}
    \label{fig:mask-filt-samp}
    \vspace{-0.5em}
\end{figure}

%% file: sections/05_results.tex
\section{Experimental Results}
\label{sec:results}

First, we evaluate the event representations and filtering approaches in object detection tasks, which is also a key component of the \textit{ hybrid} pose estimation approach. The best-performing representation and filtering technique will be adopted for the baseline evaluation of pose estimation. Sub-sections A to C present the experimental details and results for event data representations and filtering techniques. Followed by sub-sections D to F, it evaluates baseline pose estimation algorithms on the proposed SPADES dataset.

\subsection{Metrics for Object Detection}
The results of object detection tasks are evaluated using the standard metrics, Average Precision (AP) and Average Recall (AR) at varying Intersection-over-Union (IoU) thresholds 0.5, 0.75 and for object bounding box sizes \cite{padilla2020survey}: small (S) [$A_{bbox} \leq 150\times 150$], medium (M) [$150\times 150 <  A_{bbox} \leq 300\times 300$] and large (L) [$300\times 300 < A_{bbox}$], where $A_{bbox}$ denotes the area of the bounding box in sq. pixels.

\subsection{Event Representation}
\label{sub-sec:event-repr}
The image-based representations E2F, LNES, TS, and 3C are evaluated on an object detection task. A Faster-RCNN model with Mobilenet-V3-Large \cite{howard2019searching} backbone was used. The model was trained on a batch size of 8 for 100 epochs, with early stopping patience set to 20 epochs. The model's backbone was initialised using pre-trained weights on the Imagenet \cite{deng2009imagenet} dataset. \Cref{tab:eval-event-repr} show that the 3C representation yields the most favourable results for object detection in both synthetic and real data on both metrics. 

\begin{table}[ht]
    \tabcolsep=0.12cm
    \centering
    \footnotesize
    \begin{tabularx}{\columnwidth}{lccccccccc}
    \toprule
         \multirow{2}{*}{\texttt{Rep.}} & \multicolumn{9}{c}{\texttt{Metrics}} \\
          & AP$_{0.5}$ & AP$_{0.75}$ & AP$_{S}$  & AP$_{M}$ & {AP$_{L}$} & AR & AR$_{S}$ & AP$_{M}$ & AR$_{L}$ \\ 
        \midrule
         & \multicolumn{9}{c}{\texttt{Synthetic}} \\
         \midrule
        {E2F} & 0.98 & 0.74 & 0.60 & 0.51 & 0.61 & 0.66 & 0.67 & 0.61 & 0.63 \\ 
        {LNES} & 0.98 & 0.73 & 0.59 & 0.51 & 0.63 & 0.66 & 0.66 & 0.62 & 0.66 \\ 
        {TS} & 0.98 & 0.74 & 0.59 & 0.52 & 0.65 & 0.65 & 0.65 & 0.63 & 0.64 \\ 
        {3C} & \textbf{0.99} & \textbf{0.95} & \textbf{0.84} & \textbf{0.82} & \textbf{0.79} & \textbf{0.89} & \textbf{0.89} & \textbf{0.90} & \textbf{0.83} \\ 
        \midrule
         & \multicolumn{9}{c}{\texttt{Real}} \\
        \midrule
        {E2F} & 0.69 & 0.49 & \textbf{0.45} & 0.46 & 0.33 & 0.55 & 0.54 & 0.56 & 0.55 \\ 
        {LNES} & 0.63 & 0.49 & 0.43 & 0.44 & 0.22 & 0.54 & \textbf{0.59} & 0.54 & 0.34 \\ 
        {TS} & 0.63 & 0.48 & 0.42 & 0.44 & 0.27 & 0.53 & 0.57 & 0.55 & 0.38 \\ 
        {3C} & \textbf{0.71} & \textbf{0.50} & 0.40 & \textbf{0.48} & \textbf{0.38} & \textbf{0.58} & 0.55 & \textbf{0.59} & \textbf{0.57} \\        
        \bottomrule
    \end{tabularx}
    \caption{Evaluation of event representations on synthetic (test) and real datasets.}
     \label{tab:eval-event-repr}
    \vspace{-2em}
\end{table}

%


\subsection{Filtering}
To find the best filtering approach, two additional datasets were derived using the event filtering methods presented in \cref{sub-sec:data-filt}. The full synthetic data set (without filtering) has 179,700 pose labels; the mask-based filtering resulted in 94,147 labels (52\%), and the bbox-based filtering resulted in 113,876 labels (63\%) for training. Three CNN models having a similar architecture as in \cref{sub-sec:event-repr} were trained and validated on the \textit{real dataset}. All models used the \textit{3C representation} for the training and evaluation. It is important to note that the real event data is filtered only using a simple event count-based filtering ($>$10,000 events) to select the valid event frame. The results summarized in \Cref{tab:eval-event-ftr} show that the \textit{mask-based filtering} outperforms other methods in real data even when trained with a lower number of good quality data. Based on the results, we adopt \textit{3C representation with mask filtering} as a preprocessing step to filter training data for the baseline evaluation of pose estimation.  

\begin{table}[ht]
    \vspace{0.5em}
    \centering
    \footnotesize
    \tabcolsep=0.105cm
    \begin{tabularx}{\linewidth}{lccccccccc}
    \toprule
        \multirow{2}{*}{\texttt{Filter}} & \multicolumn{9}{c}{\texttt{Metrics}} \\
          & AP$_{0.5}$ & AP$_{0.75}$ & AP$_S$  & AP$_M$ & {AP$_L$} & AR & AR$_S$ & AP$_M$ & AR$_L$ \\ 
        \midrule
        & \multicolumn{9}{c}{\texttt{Synthetic}} \\
         \midrule
        {w/o Filt.} & \textbf{0.99} & \textbf{0.95} & \textbf{0.84} & \textbf{0.82} & \textbf{0.79} & \textbf{0.89} & \textbf{0.89} & \textbf{0.90} & \textbf{0.83} \\[0.1cm]
        {Bbox} & 0.89 & 0.58 & 0.69 & 0.52 & 0.67 & 0.75 & 0.77 & 0.68 & 0.78 \\[0.1cm]
        {Mask} & 0.98 & 0.84 & 0.74 & 0.59 & 0.66 & 0.79 & 0.80 & 0.70 & 0.81 \\
        \midrule
         & \multicolumn{9}{c}{\texttt{Real}} \\
        \midrule
        {w/o Filt.} & 0.69 & 0.48 & 0.38 & 0.49 & 0.38 & 0.54 & 0.52 & 0.59 & 0.57 \\[0.1cm]
        {Bbox} & 0.71 & 0.50 & 0.40 & 0.48 & 0.38 & 0.58 & 0.55 & 0.59 & 0.57 \\[0.1cm]
        {Mask} & \textbf{0.72} & \textbf{0.53} & \textbf{0.43} & \textbf{0.51} & \textbf{0.41} & \textbf{0.59} & \textbf{0.59} & \textbf{0.61} & \textbf{0.57} \\ \bottomrule
    \end{tabularx}
    \caption{Evaluation of filtering techniques on synthetic (test) and real datasets.}   
    \label{tab:eval-event-ftr}
    \vspace{-2em}
\end{table}

\subsection{Pose Estimation Baseline Algorithms}
\label{sec:alg_val}
The baseline study explores two prominent algorithmic approaches in spacecraft pose estimation: \textit{Direct} approach and \textit{Hybrid} approach. A two-branch network strategy was adopted for the \textit{Direct} approach. The first branch is dedicated to translation prediction, and it employs a CNN backbone on the entire image, followed by a fully connected layer. The second branch is for rotation prediction, employing a CNN backbone on the Region of Interest (RoI) extracted from an object detector, followed by a fully connected layer to regress rotation parameters represented by the Fisher Matrix Representation \cite{lee2018bayesian}. In the \textit{Hybrid} approach, network architecture is similar to \cite{rathinam2020}, where we utilized Faster-RCNN with ResNet-50 backbone for object detection, HigherHRNet for keypoint regression and BPnP \cite{chen2020end} for PnP optimization during inference.

\subsection{Pose Estimation Metrics}
 The pose error metrics defined in \cite{park2021} were used to analyse the results, except for translation error, which is changed to a relative translation error. The metrics are the relative translation error $E_\textrm{T}$ [\%], rotation error $E_\textrm{R}$ [$^\circ$], and pose error $E_\textrm{P}$ defined as below. 
\begin{equation*}
   E_\textrm{T} = \frac{\| \tilde{\bm{t}} - \bm{t} \|_2}{\| \bm{t} \|_2}; \; \; \; \; E_\textrm{R} = 2\ acos |\innerproduct{\tilde{q}}{q}|; 
   \; \; \; \; E_\textrm{P} = E_\textrm{R} + E_\textrm{T}.  
\end{equation*}

\subsection{Pose Estimation Performance}

The baseline results for the \textit{Direct} and \textit{Hybrid} approach evaluated on synthetic and real datasets are summarized in \cref{tab:results_eval}. The metric labeled \textit{ Data [\%]} indicates the proportion of data that the algorithm could confidently infer a pose based on a threshold. Two confidence thresholds were employed within the \textit{Hybrid} approach: 0.9 for object detection and 0.5 for keypoint regression. Consequently, $>75$\% of the synthetic test dataset was filtered out. Although the \textit{Hybrid} approach exhibits superior performance, it yielded results on a substantially lower percentage of data than the \textit{Direct} approach. This was primarily due to the fact that keypoint predictions frequently fell below the threshold and there were insufficient keypoints for PnP to estimate the pose. 

\begin{table}[ht]
	\centering
	\footnotesize
	\tabcolsep=0.19cm
	\begin{tabularx}{\linewidth}{l|cccc|cccc}
		\toprule
		\multirow{2}{*}{\texttt{Model}} & Data & $E_\textrm{T} $ & $E_\textrm{R}$  & $E_\textrm{P}$ & Data & $E_\textrm{T} $ & $E_\textrm{R}$  & $E_\textrm{P}$ \\
  		 & [\%] & [\%] & [$^\circ$] & [-] & [\%] & [\%] &  [$^\circ$] &  [-] \\
		\midrule
  		& \multicolumn{4}{c|}{\texttt{Synthetic}} & \multicolumn{4}{c}{\texttt{Real}}  \\
    	\midrule
		Direct  & \textbf{97.32} & 4.29 & 30.43 & 0.57 & \textbf{73.32} & 5.13 & 81.13 &  1.47 \\[0.1cm]
  		Hybrid & 23.98 & \textbf{3.23} & \textbf{6.69} & \textbf{0.15} & 17.27 & \textbf{3.34} & \textbf{78.98} & \textbf{1.41}  \\
		\bottomrule
	\end{tabularx}
        \caption{Performance of baseline models on synthetic (test) and real datasets}
 	\label{tab:results_eval}
    \vspace{-1em}
\end{table}

\begin{table}[ht]
    \vspace{0.5em}
	\centering
	\footnotesize
	\tabcolsep=0.19cm
	\begin{tabularx}{\linewidth}{l|cccc|cccc}
		\toprule
		\multirow{2}{*}{\texttt{Model}} & Data & $E_\textrm{T} $ & $E_\textrm{R}$  & $E_\textrm{P}$ & Data & $E_\textrm{T} $ & $E_\textrm{R}$  & $E_\textrm{P}$ \\
  		 & [\%] & [\%] & [$^\circ$] & [-] & [\%] & [\%] &  [$^\circ$] &  [-] \\
		\midrule
  		& \multicolumn{4}{c|}{\texttt{No-BG + Easy-LI}}  & \multicolumn{4}{c}{\texttt{BG + Easy-LI}} \\
    \midrule
		Direct  & \textbf{99.89} & 2.21 & 19.99 & 0.37 & \textbf{98.78} & \textbf{2.93} & 29.14 & 0.54  \\[0.1cm]
		Hybrid & 25.87 & \textbf{2.12} & \textbf{2.47} & \textbf{0.06} & 24.57 & 2.98 & \textbf{3.76} & \textbf{0.09}  \\
		\midrule 
		& \multicolumn{4}{c|}{\texttt{No-BG + Hard-LI}} & \multicolumn{4}{c}{\texttt{BG + Hard-LI}}  \\
        \midrule
		Direct  & \textbf{97.48} & 4.88 & 32.45 & 0.62 & \textbf{91.74} & 5.03 & 40.14 & 0.75  \\[0.1cm]
		Hybrid & 22.78 & \textbf{4.02} & \textbf{7.53} & \textbf{0.17} & 20.64 & \textbf{4.89} & \textbf{13.07} & \textbf{0.28}  \\
		\bottomrule
	\end{tabularx}
  \caption{Impact of background (bg) and lighting (li) conditions on synthetic (test) dataset}
 	\label{tab:eval-deep-e2e-hyb}
  \vspace{-2em}
\end{table}


Further analysis of the synthetic dataset is presented in \Cref{tab:eval-deep-e2e-hyb}. It shows that lighting conditions and the background can significantly influence the performance of event data. The results presented in \Cref{tab:results_eval} are limited to DL approaches without DA techniques and suggest a notable domain gap between synthetic and real domains present in event data, leading to a discernible performance reduction.

\begin{figure}[ht]
     \centering
     \footnotesize
     \begin{minipage}[b]{0.49\linewidth}
         \centering
         \includegraphics[width=\textwidth]{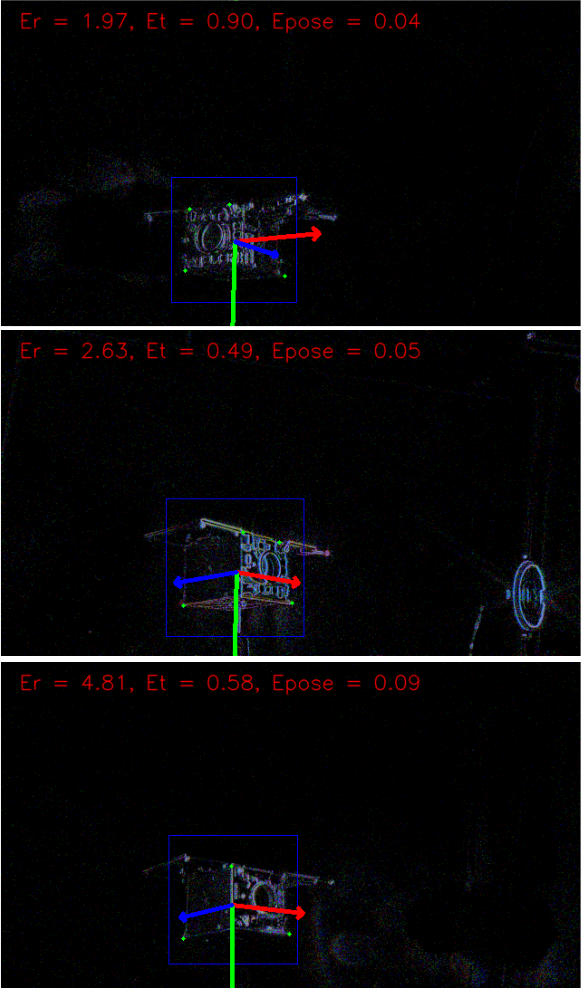}
         ( a )
     \end{minipage}
    \begin{minipage}[b]{0.49\linewidth}
         \centering
         \includegraphics[width=\textwidth]{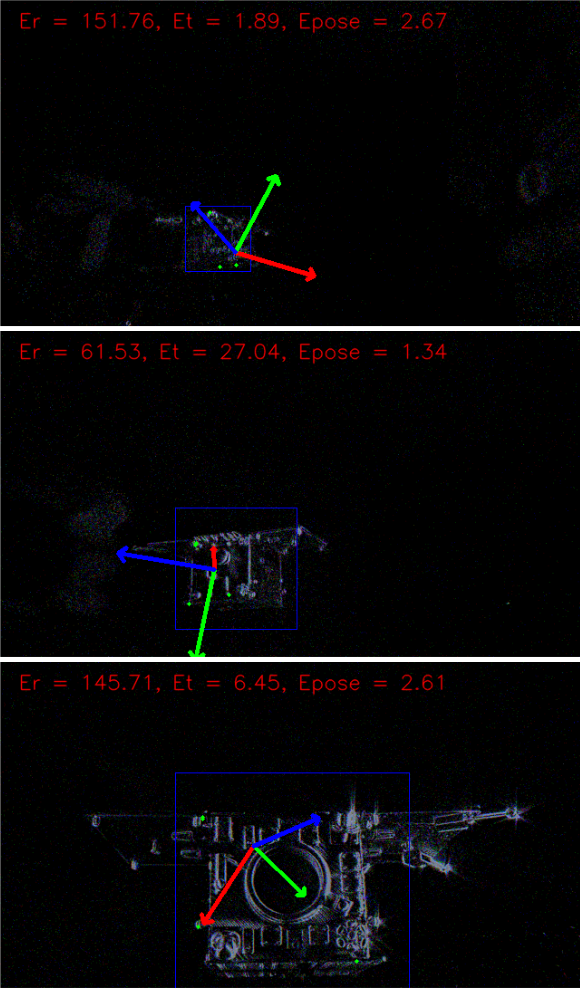}
         ( b )
     \end{minipage}
        \caption{Real dataset samples with estimated keypoints and poses. (a) Best performing, (b) Worst performing.}
        \label{fig:real-results-sample}
        \vspace{-2em}
\end{figure}

%% file: sections/06_conclusion.tex
\section{Conclusion and Future Work}
\label{sec:conclusion}

In conclusion, this work introduced the SPADES dataset, a comprehensive resource that encompasses both synthetic and real event data, designed to facilitate the training and validation of DL algorithms for event-based spacecraft pose estimation. The proposed 3-Channel event representation demonstrated superior performance compared to existing representations in object detection tasks. Furthermore, the mask-based data filtering approach improved the quality of the training data, leading to improved algorithm performance. However, the experimental results of the baseline models underscore the persistent domain gap between the synthetic and real event data.

Material properties and textures played a vital role in influencing the number of events generated in the synthetic data compared to the real data. The potential of event cameras was particularly evident in low-light conditions, where traditional RGB cameras struggled to capture useful information. In the future, we aim to refine the synthetic dataset to effectively bridge the performance gap. Additionally, leveraging the asynchronous nature of event data holds promise for advancing pose estimation and tracking techniques. 

%% file: root.bbl
\begin{thebibliography}{10}
\providecommand{\url}[1]{#1}
\csname url@rmstyle\endcsname
\providecommand{\newblock}{\relax}
\providecommand{\bibinfo}[2]{#2}
\providecommand\BIBentrySTDinterwordspacing{\spaceskip=0pt\relax}
\providecommand\BIBentryALTinterwordstretchfactor{4}
\providecommand\BIBentryALTinterwordspacing{\spaceskip=\fontdimen2\font plus
\BIBentryALTinterwordstretchfactor\fontdimen3\font minus
  \fontdimen4\font\relax}
\providecommand\BIBforeignlanguage[2]{{%
\expandafter\ifx\csname l@#1\endcsname\relax
\typeout{** WARNING: IEEEtran.bst: No hyphenation pattern has been}%
\typeout{** loaded for the language `#1'. Using the pattern for}%
\typeout{** the default language instead.}%
\else
\language=\csname l@#1\endcsname
\fi
#2}}

\bibitem{sharma2018pose}
S.~Sharma, C.~Beierle, and S.~D'Amico, ``Pose estimation for non-cooperative
  spacecraft rendezvous using convolutional neural networks,'' in \emph{2018
  IEEE Aerospace Conference}.\hskip 1em plus 0.5em minus 0.4em\relax IEEE,
  2018, pp. 1--12.

\bibitem{proenca2020}
P.~F. Proen{\c{c}}a and Y.~Gao, ``Deep learning for spacecraft pose estimation
  from photorealistic rendering,'' in \emph{2020 IEEE International Conference
  on Robotics and Automation (ICRA)}.\hskip 1em plus 0.5em minus 0.4em\relax
  IEEE, 2020, pp. 6007--6013.

\bibitem{chen2019}
B.~Chen, J.~Cao, A.~Parra, and T.-J. Chin, ``Satellite pose estimation with
  deep landmark regression and nonlinear pose refinement,'' in
  \emph{Proceedings of the IEEE/CVF International Conference on Computer Vision
  Workshops}, 2019.

\bibitem{park2022speed+}
T.~H. Park, M.~Martens, G.~Lecuyer, D.~Izzo, and S.~D'Amico, ``{{SPEED}}+:
  {{Next-Generation Dataset}} for {{Spacecraft Pose Estimation}} across
  {{Domain Gap}},'' in \emph{2022 {{IEEE Aerospace Conference}}
  ({{AERO}})}.\hskip 1em plus 0.5em minus 0.4em\relax {Big Sky, MT, USA}:
  {IEEE}, 3 2022, pp. 1--15.

\bibitem{park2021}
\BIBentryALTinterwordspacing
T.~H. Park, M.~Märtens, M.~Jawaid, Z.~Wang, B.~Chen, T.-J. Chin, D.~Izzo, and
  S.~D’Amico, ``Satellite pose estimation competition 2021: Results and
  analyses,'' \emph{Acta Astronautica}, vol. 204, pp. 640--665, 2023. [Online].
  Available:
  \url{https://www.sciencedirect.com/science/article/pii/S0094576523000048}
\BIBentrySTDinterwordspacing

\bibitem{jawaid2023}
M.~Jawaid, E.~Elms, Y.~Latif, and T.-J. Chin, ``Towards {{Bridging}} the
  {{Space Domain Gap}} for {{Satellite Pose Estimation}} using {{Event
  Sensing}},'' in \emph{2023 {{IEEE International Conference}} on {{Robotics}}
  and {{Automation}} ({{ICRA}})}.\hskip 1em plus 0.5em minus 0.4em\relax
  {London, United Kingdom}: {IEEE}, May 2023, pp. 11\,866--11\,873.

\bibitem{cohen2019event}
G.~Cohen, S.~Afshar, B.~Morreale, T.~Bessell, A.~Wabnitz, M.~Rutten, and A.~van
  Schaik, ``Event-based sensing for space situational awareness,'' \emph{The
  Journal of the Astronautical Sciences}, vol.~66, pp. 125--141, 2019.

\bibitem{afshar2020}
S.~Afshar, A.~P. Nicholson, A.~van Schaik, and G.~Cohen, ``Event-based object
  detection and tracking for space situational awareness,'' \emph{IEEE Sensors
  Journal}, vol.~20, no.~24, pp. 15\,117--15\,132, 2020.

\bibitem{gallego2022}
G.~Gallego, T.~Delbruck, G.~Orchard, C.~Bartolozzi, B.~Taba, A.~Censi,
  S.~Leutenegger, A.~J. Davison, J.~Conradt, K.~Daniilidis, and D.~Scaramuzza,
  ``Event-{{Based Vision}}: {{A Survey}},'' \emph{IEEE Transactions on Pattern
  Analysis and Machine Intelligence}, vol.~44, no.~1, pp. 154--180, Jan. 2022.

\bibitem{elms2022seenic}
\BIBentryALTinterwordspacing
E.~Elms, M.~Jawaid, Y.~Latif, and T.-J. Chin, ``{SEENIC: dataset for Spacecraft
  posE Estimation with NeuromorphIC vision},'' Oct. 2022. [Online]. Available:
  \url{https://doi.org/10.5281/zenodo.7214231}
\BIBentrySTDinterwordspacing

\bibitem{dlrproba2}
\BIBentryALTinterwordspacing
K.~Gantois, F.~Teston, O.~Montenbruck, P.~Vuilleumier, and P.~van Braembusche,
  ``Proba-2 mission and new technologies overview,'' in \emph{Small Satellite
  Systems and Services - The 4S Symposium}, December 2006. [Online]. Available:
  \url{https://elib.dlr.de/46830/}
\BIBentrySTDinterwordspacing

\bibitem{pauly2022lessons}
L.~Pauly, M.~L. Jamrozik, M.~O. Del~Castillo, O.~Borgue, I.~P. Singh, M.~R.
  Makhdoomi, O.-O. Christidi-Loumpasefski, V.~Gaudilliere, C.~Martinez,
  A.~Rathinam, \emph{et~al.}, ``{Lessons from a Space Lab--An Image Acquisition
  Perspective},'' \emph{International Journal of Aerospace Engineering}, 2023.

\bibitem{speed2019}
\BIBentryALTinterwordspacing
S.~Sharma, T.~H. Park, and S.~D'Amico, ``Spacecraft {Pose} {Estimation}
  {Dataset} ({SPEED}),'' 2019. [Online]. Available:
  \url{https://purl.stanford.edu/dz692fn7184}
\BIBentrySTDinterwordspacing

\bibitem{rathinam2022dataset}
\BIBentryALTinterwordspacing
A.~Rathinam, V.~Gaudilliere, M.~A. Mohamed~Ali, M.~Ortiz Del~Castillo,
  L.~Pauly, and D.~Aouada, ``{SPARK 2022 Dataset: Spacecraft Detection and
  Trajectory Estimation},'' June 2022. [Online]. Available:
  \url{https://doi.org/10.5281/zenodo.6599762}
\BIBentrySTDinterwordspacing

\bibitem{park2022shirt}
T.~H. Park and S.~D'Amico, ``{{SHIRT}}: {{Satellite Hardware-In-the-loop
  Rendezvous Trajectories Dataset}},'' 2022.

\bibitem{rathinam2021}
\BIBentryALTinterwordspacing
A.~Rathinam, Z.~Hao, and Y.~Gao, ``Autonomous visual navigation for spacecraft
  on-orbit operations,'' in \emph{Space {Robotics} and {Autonomous} {Systems}:
  {Technologies}, advances and applications}.\hskip 1em plus 0.5em minus
  0.4em\relax Institution of Engineering and Technology, 8 2021, pp. 125--157.
  [Online]. Available: \url{https://doi.org/10.1049/PBCE131E_ch5}
\BIBentrySTDinterwordspacing

\bibitem{pauly2023survey}
L.~Pauly, W.~Rharbaoui, C.~Shneider, A.~Rathinam, V.~Gaudillière, and
  D.~Aouada, ``A survey on deep learning-based monocular spacecraft pose
  estimation: Current state, limitations and prospects,'' \emph{Acta
  Astronautica}, vol. 212, pp. 339--360, 2023.

\bibitem{rudnev2020}
V.~Rudnev, V.~Golyanik, J.~Wang, H.-P. Seidel, F.~Mueller, M.~A. Elgharib, and
  C.~Theobalt, ``{{EventHands}}: {{Real-time}} neural {{3D}} hand pose
  estimation from an event stream,'' \emph{2021 IEEE/CVF International
  Conference on Computer Vision (ICCV)}, pp. 12\,365--12\,375, 2020.

\bibitem{lagroce2017}
X.~Lagorce, G.~Orchard, F.~Galluppi, B.~E. Shi, and R.~B. Benosman, ``Hots: A
  hierarchy of event-based time-surfaces for pattern recognition,'' \emph{IEEE
  Transactions on Pattern Analysis and Machine Intelligence}, vol.~39, no.~7,
  pp. 1346--1359, 2017.

\bibitem{xie2022voxel}
B.~Xie, Y.~Deng, Z.~Shao, H.~Liu, and Y.~Li, ``Vmv-gcn: Volumetric multi-view
  based graph cnn for event stream classification,'' \emph{IEEE Robotics and
  Automation Letters}, vol.~7, no.~2, pp. 1976--1983, 2022.

\bibitem{bi2020graph}
Y.~Bi, A.~Chadha, A.~Abbas, E.~Bourtsoulatze, and Y.~Andreopoulos,
  ``Graph-based spatio-temporal feature learning for neuromorphic vision
  sensing,'' \emph{IEEE Transactions on Image Processing}, vol.~29, pp.
  9084--9098, 2020.

\bibitem{sekikawa2019eventnet}
Y.~Sekikawa, K.~Hara, and H.~Saito, ``Eventnet: Asynchronous recursive event
  processing,'' in \emph{Proceedings of the IEEE/CVF Conference on Computer
  Vision and Pattern Recognition}, 2019, pp. 3887--3896.

\bibitem{gallego2017acc}
G.~Gallego and D.~Scaramuzza, ``Accurate angular velocity estimation with an
  event camera,'' \emph{IEEE Robotics and Automation Letters}, vol.~2, no.~2,
  pp. 632--639, 2017.

\bibitem{qiu2016}
W.~Qiu and A.~Yuille, ``{{UnrealCV}}: {{Connecting Computer Vision}} to
  {{Unreal Engine}},'' in \emph{Computer {{Vision}} \textendash{} {{ECCV}} 2016
  {{Workshops}}}, G.~Hua and H.~J{\'e}gou, Eds.\hskip 1em plus 0.5em minus
  0.4em\relax {Cham}: {Springer International Publishing}, 2016, vol. 9915, pp.
  909--916.

\bibitem{joubert2021}
D.~Joubert, A.~Marcireau, N.~Ralph, A.~Jolley, A.~Van~Schaik, and G.~Cohen,
  ``Event {{Camera Simulator Improvements}} via {{Characterized Parameters}},''
  \emph{Frontiers in Neuroscience}, vol.~15, p. 702765, July 2021.

\bibitem{prophesee2023}
\BIBentryALTinterwordspacing
``\BIBforeignlanguage{en-US}{Event {Camera} {Evaluation} {Kit} 4 {HD} {IMX636}
  {Prophesee}-{Sony}}.'' [Online]. Available:
  \url{https://www.prophesee.ai/event-camera-evk4/}
\BIBentrySTDinterwordspacing

\bibitem{muglikar2021}
\BIBentryALTinterwordspacing
M.~Muglikar, M.~Gehrig, D.~Gehrig, and D.~Scaramuzza, ``How to calibrate your
  event camera,'' 2021. [Online]. Available:
  \url{https://arxiv.org/abs/2105.12362}
\BIBentrySTDinterwordspacing

\bibitem{Rebecq19pami}
\BIBentryALTinterwordspacing
H.~Rebecq, R.~Ranftl, V.~Koltun, and D.~Scaramuzza, ``High speed and high
  dynamic range video with an event camera,'' \emph{{IEEE} Trans. Pattern Anal.
  Mach. Intell. (T-PAMI)}, 2019. [Online]. Available:
  \url{http://rpg.ifi.uzh.ch/docs/TPAMI19_Rebecq.pdf}
\BIBentrySTDinterwordspacing

\bibitem{radu_handeye_1995}
R.~Horaud and F.~Dornaika, ``Hand-eye calibration,'' \emph{The International
  Journal of Robotics Research}, vol.~14, no.~3, pp. 195--210, 1995.

\bibitem{bai_accurate_2022}
\BIBentryALTinterwordspacing
W.~Bai, Y.~Chen, R.~Feng, and Y.~Zheng, ``Accurate and {Efficient}
  {Frame}-based {Event} {Representation} for {AER} {Object} {Recognition},'' in
  \emph{2022 {International} {Joint} {Conference} on {Neural} {Networks}
  ({IJCNN})}.\hskip 1em plus 0.5em minus 0.4em\relax Padua, Italy: IEEE, July
  2022, pp. 1--6. [Online]. Available:
  \url{https://ieeexplore.ieee.org/document/9892070/}
\BIBentrySTDinterwordspacing

\bibitem{padilla2020survey}
R.~Padilla, S.~L. Netto, and E.~A. Da~Silva, ``A survey on performance metrics
  for object-detection algorithms,'' in \emph{2020 international conference on
  systems, signals and image processing (IWSSIP)}.\hskip 1em plus 0.5em minus
  0.4em\relax IEEE, 2020, pp. 237--242.

\bibitem{howard2019searching}
A.~Howard, M.~Sandler, B.~Chen, W.~Wang, L.-C. Chen, M.~Tan, G.~Chu,
  V.~Vasudevan, Y.~Zhu, R.~Pang, \emph{et~al.}, ``{Searching for
  MobileNetV3},'' in \emph{2019 IEEE/CVF International Conference on Computer
  Vision (ICCV)}.\hskip 1em plus 0.5em minus 0.4em\relax IEEE, 2019.

\bibitem{deng2009imagenet}
J.~Deng, W.~Dong, R.~Socher, L.-J. Li, K.~Li, and L.~Fei-Fei, ``Imagenet: A
  large-scale hierarchical image database,'' in \emph{2009 IEEE conference on
  computer vision and pattern recognition}.\hskip 1em plus 0.5em minus
  0.4em\relax Ieee, 2009, pp. 248--255.

\bibitem{lee2018bayesian}
T.~Lee, ``Bayesian attitude estimation with the matrix fisher distribution on
  so (3),'' \emph{IEEE Transactions on Automatic Control}, vol.~63, no.~10, pp.
  3377--3392, 2018.

\bibitem{rathinam2020}
A.~Rathinam and Y.~Gao, ``On-{{Orbit Relative Navigation Near}} a {{Known
  Target Using Monocular Vision}} and {{Convolutional Neural Networks}} for
  {{Pose Estimation}},'' in \emph{International {{Symposium}} on {{Artificial
  Intelligence}}, {{Robotics}} and {{Automation}} in {{Space}} ({{iSAIRAS}})},
  Online, 10 2020.

\bibitem{chen2020end}
B.~Chen, A.~Parra, J.~Cao, N.~Li, and T.-J. Chin, ``End-to-end learnable
  geometric vision by backpropagating {PnP} optimization,'' in
  \emph{Proceedings of the IEEE/CVF Conference on Computer Vision and Pattern
  Recognition}, 2020, pp. 8100--8109.

\end{thebibliography}
